# MF2Summ: Multimodal Fusion for Video Summarization with Temporal Alignment

Shuo Wang[1*], Jiahao Zhang[2†]

[1]Nanjing University
[2]Huazhong University of Science and Technology
522024320155@smail.nju.edu.cn, m202476982@hust.edu.cn

## Abstract

The rapid proliferation of online video content necessitates effective video summarization techniques. Traditional methods, often relying on a single modality (typically visual), struggle to capture the full semantic richness of videos. This paper introduces MF2Summ, a novel video summarization model based on multimodal content understanding, integrating both visual and auditory information. MF2Summ employs a five-stage process: feature extraction, cross-modal attention interaction, feature fusion, segment prediction, and key shot selection. Visual features are extracted using a pre-trained GoogLeNet model, while auditory features are derived using SoundNet. The core of our fusion mechanism involves a cross-modal Transformer and an alignment-guided self-attention Transformer, designed to effectively model inter-modal dependencies and temporal correspondences. Segment importance, location, and center-ness are predicted, followed by key shot selection using Non-Maximum Suppression (NMS) and the Kernel Temporal Segmentation (KTS) algorithm. Experimental results on the SumMe and TVSum datasets demonstrate that MF2Summ achieves competitive performance, notably improving F1-scores by 1.9% and 0.6% respectively over the DSNet model, and performing favorably against other state-of-the-art methods.

## 1 Introduction

The digital era has witnessed an unprecedented explosion in video content.[Zhu et al., 2020] Platforms like YouTube receive hundreds of hours of video uploads per minute, and short-form video applications such as TikTok boast billions of users with extensive daily engagement. This deluge of information makes it increasingly challenging for users to efficiently find and consume relevant video content. Video summarization technology addresses this challenge by aiming to extract the most valuable information from a video, generating a concise representation that either encapsulates the main content or highlights the most compelling segments. [Zhu et al., 2020]2 The general pipeline for video summarization typically involves three main steps: 1) shot boundary detection to segment the video into coherent units; 2) frame-level or shot-level importance score prediction to identify salient parts; and 3) key shot selection to compile the final summary. However, many traditional and even some contemporary approaches rely predominantly on visual information. This unimodal focus can be a significant limitation, as visual features alone may not capture the complete contextual and semantic nuances of a video. For instance, crucial information might be conveyed through dialogue, sound events, or the emotional tone of background music, all of which are lost in a purely visual analysis. [Psallidas et al., 2021] To overcome these limitations, multimodal video summarization has emerged as a promising research direction. [Jangra et al., 2023]By integrating information from various modalities—such as visual streams, audio tracks, and textual data (e.g., subtitles or user comments)—these methods aim to achieve a more comprehensive understanding of video content. The synergy between different modalities can enhance the accuracy of importance prediction, reduce information redundancy, and lead to more diverse and engaging summaries. Audio, in particular, is an inherently synchronized modality with video and often carries significant semantic cues, including spoken language, environmental sounds, and emotional undertones conveyed by music, which are complementary to visual information. [Baltrušaitis et al., 2018] This paper proposes MF2Summ, a novel video summarization network that leverages both visual and auditory modalities. The core premise of MF2Summ is that by effectively fusing features from these two modalities through sophisticated attention mechanisms, we can generate more accurate and representative video summaries. Specifically, MF2Summ utilizes a cross-modal Transformer to facilitate interaction between visual and auditory features and an alignment-guided self-attention Transformer to perform a temporally aware fusion of these features. The main contributions of this work are:

- A novel multimodal video summarization network, MF2Summ, that effectively integrates visual features extracted by GoogLeNet [Szegedy et al., 2015] and auditory features (including emotional cues from sound) extracted by SoundNet [Aytar et al., 2016]

---

[*]Equal contribution.
[†]Equal contribution.

- A sophisticated fusion mechanism based on a cross-modal Transformer architecture, inspired by models like MulT [Tsai *et al.*, 2019], combined with an alignment-guided self-attention Transformer. This dual-transformer approach is designed to capture intricate inter-modal dependencies and leverage temporal correspondences between visual and auditory streams

- A comprehensive experimental evaluation on two standard benchmark datasets, SumMe and TVSum [Zhang *et al.*, 2016a]. The results demonstrate that MF2Summ achieves competitive performance against existing state-of-the-art methods and significantly outperforms strong visual-only baselines like DSNet [Zhu *et al.*, 2020].

## 2 Related Work

Video summarization research has explored various paradigms, broadly categorized by the level of supervision required and the modalities utilized.

### 2.1 Supervised Video Summarization

Supervised methods typically rely on datasets with frame-level or shot-level importance scores annotated by humans. Early deep learning approaches often employed Long Short-Term Memory (LSTM) networks to model temporal dependencies among video frames. For example, Zhang et al. introduced an LSTM-based model for selecting keyframes or key sub-shots [Zhang *et al.*, 2016a], and similar LSTM architectures like vsLSTM and dppLSTM have been explored [Zhang *et al.*, 2016b]. Other works have utilized 3D Convolutional Neural Networks (3D CNNs) to capture spatio-temporal features directly from video segments. More recently, attention mechanisms have gained prominence. Fajtl et al. proposed VASNet, a self-attention based network for video summarization [Fajtl *et al.*, 2019]. While these methods can achieve good performance by learning from human judgments, they are often constrained by the high cost and subjective nature of manual annotation.

### 2.2 Unsupervised and Weakly-Supervised Video Summarization

To alleviate the dependency on extensive annotations, unsupervised and weakly-supervised methods have been developed. Unsupervised approaches aim to identify important segments without any labeled data, often relying on criteria like representativeness or diversity. Generative Adversarial Networks (GANs) have been popular in this domain, with models like SUM-GAN [Mahasseni *et al.*, 2017] learning to generate summaries that are difficult to distinguish from original video characteristics. Reinforcement learning has also been applied, for instance, DR-DSN [Zhou *et al.*, 2018] uses a diversity-representativeness reward to train a summarization network. Weakly-supervised methods leverage auxiliary information, such as video titles or categories, which are easier to obtain than dense frame-level scores [Panda *et al.*, 2017]. These approaches offer greater scalability but may not always capture fine-grained user preferences.

### 2.3 Multimodal Video Summarization

Recognizing the limitations of unimodal approaches, research has increasingly focused on multimodal video summarization [Jangra *et al.*, 2023]. The integration of information from different sources, such as visual content, audio, and text, can lead to a richer understanding of the video.

A significant body of work has explored visual-textual fusion. CLIP-It [Narasimhan *et al.*, 2021], for example, leverages the power of CLIP embeddings and uses a language-guided Transformer to score frames based on their relevance to either a user query or automatically generated captions. While effective, such methods depend on the availability and quality of textual information. The Multimodal Transformer (MulT), proposed by Tsai et al. [Tsai *et al.*, 2019], provides a foundational architecture for handling unaligned multimodal sequences using cross-modal attention. This allows different modalities to attend to each other, facilitating information exchange and fusion. Our proposed MF2Summ model builds upon these concepts of cross-modal attention and alignment but focuses on the audio-visual domain.

### 2.4 Positioning MF2Summ

MF2Summ distinguishes itself from prior work in several ways. Compared to unimodal supervised methods like DSNet [Zhu *et al.*, 2020] (which formulates summarization as a temporal interest detection problem), MF2Summ incorporates the auditory modality and a more sophisticated fusion strategy. While CLIP-It [Narasimhan *et al.*, 2021] also employs multimodal fusion, it focuses on language guidance. MF2Summ, in contrast, explores the synergy between visual and auditory cues, which are naturally synchronized in videos. Audio can provide direct information about events, speech, and emotional context without relying on an intermediate text generation step. The novelty of MF2Summ lies in its specific audio-visual fusion architecture, which combines a cross-modal Transformer for inter-modality adaptation with an alignment-guided self-attention Transformer for robust, temporally coherent fusion.

## 3 Preliminary

In this part, we briefly introduce the key technologies that form the foundation of our proposed MF2Summ model.

### 3.1 Feature Representation

**GoogLeNet for Visual Features** GoogLeNet, also known as Inception-v1, is a deep convolutional neural network architecture renowned for its effectiveness in image classification and feature extraction [Chen *et al.*, 2019]. Its core innovation is the "Inception module," which uses parallel convolutional filters of different sizes (e.g., 1x1, 3x3, 5x5) at the same level. This multi-scale processing allows the network to capture visual patterns at various resolutions simultaneously. By stacking these modules, GoogLeNet builds a rich hierarchical representation of image content. In our work, we use a pre-trained GoogLeNet (with its final classification layers removed) as a robust backbone to extract high-level spatial features from each video frame.

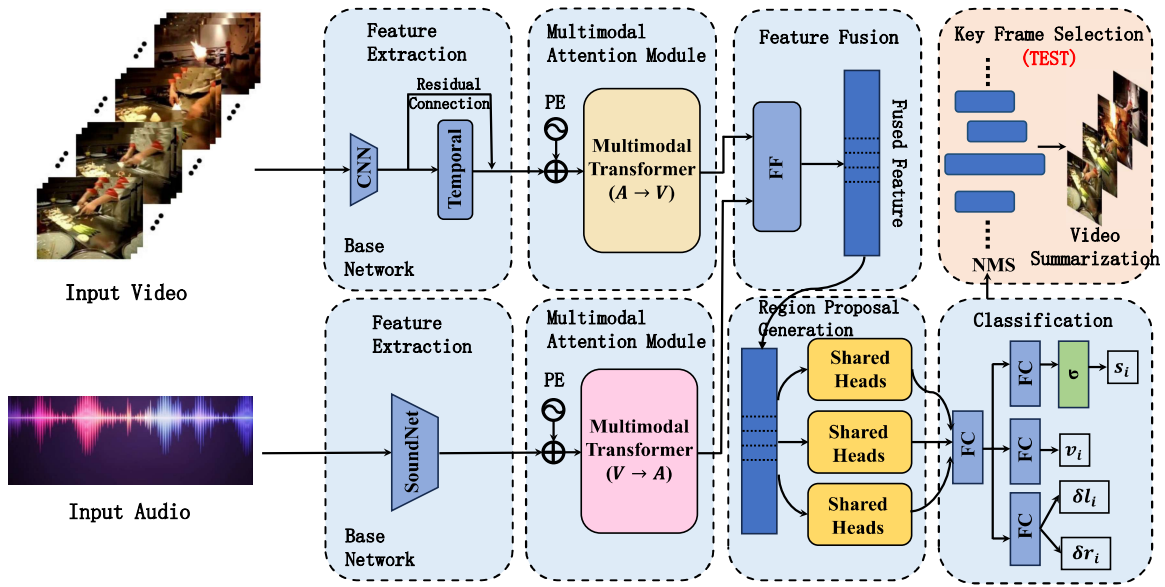

Figure 1: Overall architecture of the MF2Summ model

**SoundNet for Auditory Features** SoundNet is a deep convolutional network designed to learn sound representations directly from unlabeled video data [Aytar et al., 2016]. It leverages the natural synchronization between visual and auditory events in videos. By training the network to predict visual information (e.g., objects and scenes recognized by pre-trained visual models) from the raw audio waveform, SoundNet learns a rich auditory feature space. These features capture semantic information about the acoustic environment, such as ambient sounds, dialogue, and music, which are highly complementary to the visual content for a comprehensive video understanding. We utilize a pre-trained SoundNet to extract auditory scene features that correspond temporally to the video frames.

### 3.2 Cross-Modal Attention

The Transformer architecture, originally proposed for machine translation, has become a cornerstone of sequence modeling, largely due to its self-attention mechanism [Vaswani et al., 2017]. A key extension of this concept for multimodal tasks is cross-modal attention. This mechanism allows one modality to attend to another, creating a contextually enriched representation. Given a target modality sequence $X_\alpha$ and a source modality sequence $X_\beta$, cross-modal attention computes an updated representation for the target by using queries $Q_\alpha$ derived from $X_\alpha$ to attend to keys $K_\beta$ and values $V_\beta$ derived from $X_\beta$. The formulation is as follows:

$$\text{CrossAttention}(X_\alpha, X_\beta) = \text{softmax}(\frac{Q_\alpha K_\beta^T}{\sqrt{d_k}})V_\beta$$

are linear projections. This operation allows elements in the target sequence to selectively draw information from the source sequence, forming the basis for effective multimodal fusion as seen in models like the Multimodal Transformer (MulT) [Tsai et al., 2019]. Our model heavily relies on this mechanism to enable meaningful interaction between the visual and auditory streams.

In this part, we detail the architecture and components of the proposed Multimodal Fusion for Video Summarization Network (MF2Summ). Our approach formulates video summarization as a segment proposal and selection problem, where the model learns to identify temporally coherent and important segments by fusing audio-visual information.

## 4 Methodology

### 4.1 Overall Architecture

The MF2Summ model is designed to process multimodal input (video frames and corresponding audio sequences) to generate a concise video summary. The architecture, depicted in Figure 1, comprises five main stages:

1. **Multimodal Feature Extraction:** Separate encoders process visual and auditory data to obtain feature representations.

2. **Cross-modal Attention Interaction:** A cross-modal Transformer facilitates information exchange between the visual and auditory feature streams.

3. **Feature Fusion:** An alignment-guided self-attention Transformer fuses the interacted multimodal features, respecting temporal correspondence.

4. **Segment Prediction:** Based on the fused features, the model predicts frame-level importance scores, segment boundaries, and center-ness scores.

5. **Key Summary Generation:** Post-processing steps, including Non-Maximum Suppression (NMS) and the Kernel Temporal Segmentation (KTS) algorithm, are used to select and assemble the final summary.

## 4.2 Multimodal Feature Extraction

**Visual Feature Extraction** To capture rich visual information, we first process each video frame to obtain its spatial content representation. As shown in Figure 2, we use a pre-trained GoogLeNet model [Chen *et al.*, 2019] with its final classification layers removed to extract a 1024-dimensional deep feature vector $v_j$ for each frame $j$.

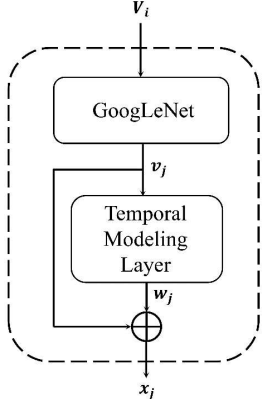

Figure 2: The architecture of our visual feature extraction module.

However, understanding a video requires modeling temporal context. Capturing long-range dependencies is crucial for identifying representative segments and understanding the overall narrative [Potapov *et al.*, 2014]. Therefore, we apply a temporal modeling layer to the sequence of frame features $v_j$. While various architectures like LSTM or GCN could be used, our default implementation employs a self-attention mechanism [Vaswani *et al.*, 2017], which has proven effective at capturing long-range dependencies. This layer computes a temporally contextualized representation $w_j$ for each frame. The final visual representation $x_j$ is then obtained by integrating the spatial and temporal features via a residual connection: $x_j = w_j + v_j$.

**Auditory Feature Extraction** To incorporate auditory information, we extract features from the raw audio waveform corresponding to the video. We utilize a pre-trained SoundNet model [Aytar *et al.*, 2016], a deep convolutional network designed to learn sound representations from unlabeled videos. The input audio is first pre-processed: converted to MP3 format, resampled to 22 kHz, and converted to a single (mono) channel. The SoundNet model then processes segments of the audio waveform that are temporally aligned with the video frames. This results in a sequence of audio feature vectors $a_j$, forming the auditory scene feature sequence.

## 4.3 Multimodal Interaction and Fusion

Effective fusion of visual and auditory features is critical for leveraging their complementary information. MF2Summ employs a two-stage process involving positional embedding, cross-modal attention, and alignment-guided self-attention.as illustrated in Figure 2??.

Transformer architectures are inherently permutation-invariant. To provide them with temporal context, we add fixed sinusoidal positional embeddings (PE) to the input feature sequences of both modalities, following the original Transformer design [Vaswani *et al.*, 2017]. For a sequence of length T and feature dimension d, the PE for position pos and dimension $i$ is calculated as:

$$PE(pos, 2i) = \sin(pos/10000^{2i/d})$$
$$PE(pos, 2i+1) = \cos(pos/10000^{2i/d})$$

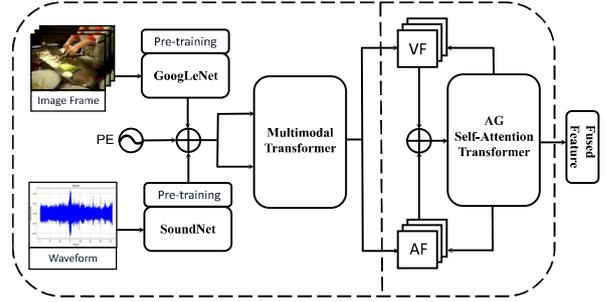

Figure 3: The overall process of multimodal feature interaction and fusion.

**Cross-Modal Transformer** Inspired by the Multimodal Transformer (MulT) [Tsai *et al.*, 2019], we employ a cross-modal Transformer to enable directed interaction between the visual and auditory modalities. This allows one modality to adapt its representation based on information from the other. The core mechanism is shown in Figure 4. For instance, in a V→A interaction, the auditory features are refined by attending to visual features. The cross-modal attention mechanism computes an attended representation for a target modality by using its features as queries ($Q_\alpha$) and the source modality's features as keys ($K_\beta$) and values ($V_\beta$). This process yields contextually enriched representations for each modality, having "listened" to the other. We apply this bidirectionally, creating both V-attended-by-A and A-attended-by-V feature streams.

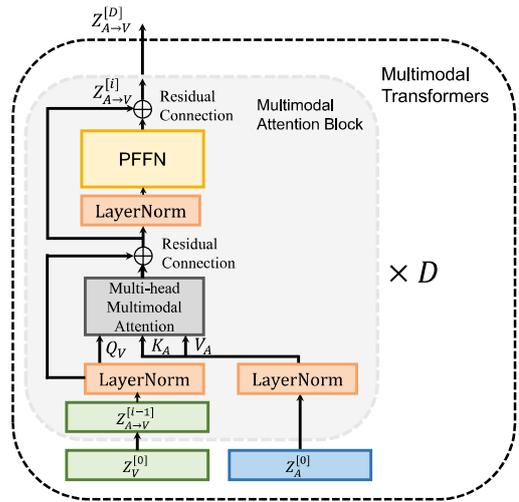

Figure 4: The architecture of the Cross-Modal Attention Block.

**Alignment-Guided Self-Attention Transformer** The outputs from the parallel cross-modal Transformer streams are concatenated. This concatenated feature sequence, now rich with cross-modal information, is fed into an Alignment-Guided Self-Attention Transformer.

Standard global self-attention applied to uncurated multimodal sequences can be suboptimal, as there can be significant amounts of irrelevant background information or imperfect synchronization. The alignment-guided self-attention module addresses this by explicitly leveraging the temporal correspondence between video and audio. It uses a masked self-attention mechanism. An attention mask A is constructed such that:

- Within the same modality (e.g., video-to-video), global attention is allowed.
- Across modalities (e.g., video-to-audio), attention is restricted to only those audio segments that are temporally co-occurrent with the video frame.

This masked attention matrix is then applied during the standard self-attention calculation by element-wise multiplication with the attention scores before the softmax operation. This alignment-guided fusion produces a unified multimodal representation that is sensitive to temporal synchrony, mitigating noise from unrelated segments and enhancing the quality of the fused features.

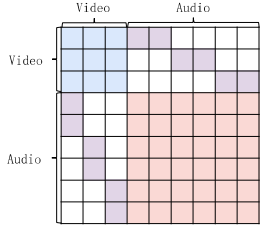

Figure 5: Diagram of the Alignment-Guided Self-Attention Mask

## 4.4 Segment Prediction and Loss Function

Using the fused multimodal representations, MF2Summ predicts three key attributes for each temporal position (frame):

1. **Importance Score** ($s_j$): A classification task to determine if frame j should be part of the summary.

2. **Segment Boundaries** ($\delta_l, \delta_r$): For each positive frame, the model regresses the temporal distances to the start ($t_{start}$) and end ($t_{end}$) of the summary segment it belongs to: $\delta_l^* = j - t_{start}$ and $\delta_r^* = t_{end} - j$.

3. **Center-ness Score** ($v_j$): This score, defined as $v_j^* = \sqrt{\frac{\min(\delta_l, \delta_r)}{\max(\delta_l, \delta_r)}}$, encourages the predicted boundaries to be centered around the frame, helping to avoid low-quality segments predicted near true segment edges.

The model is trained end-to-end using a multi-task loss function $L^*$:

$$L^* = L_{cls} + \lambda L_{reg} + \mu L_{center}$$

**Algorithm 1** MF2Summ Training Procedure

**Input**: video sequences $\{V_i\}$, audio sequences $\{A_i\}$, frame-level annotations $\{u_i^*\}$
**Output**: importance scores $s$, boundary intervals $\delta t = (\delta l, \delta r)$, centre-ness scores $v$, model parameters $\theta$

1: $\theta \leftarrow$ PARAMETERS(MFF–DSNet)
2: **for** *epoch* = 1 **to** M **do**
3:   **for** video $V \in \{V_i\}$ **and** annotation $u^* \in \{u_i^*\}$ **do**
4:     % Apply KTS and knapsack
5:     $s^* \leftarrow$ TOSHOTLEVELANNOTATION($u^*$)
6:     **for** frame $l_j \in V$ **do**
7:       $v_j \leftarrow$ GOOGLENET($l_j$)
8:       $s_j^* \leftarrow$ ASSIGNLABEL($j, s^*$)
9:       $\delta t_j^* \leftarrow$ COMPUTEBOUNDARY($j, s^*$)  % Eq. (3.7)
10:      $v_j^* \leftarrow$ COMPUTECENTERNESS($\delta t_j^*$)  % Eq. (3.9)
11:     **end for**
12:     $w \leftarrow$ EXTRACTTEMPORALFEATURE($v$)
13:     $x \leftarrow w + v$
14:     $s \leftarrow$ CLASSIFICATIONBRANCH($x$)   % importance scores
15:     $v \leftarrow$ CENTERNESSBRANCH($x$)    % centre-ness scores
16:     $\delta t \leftarrow$ REGRESSIONBRANCH($x$)     % boundary intervals
17:     $\mathcal{L} \leftarrow \mathcal{L}(s, s^*, \delta t, \delta t^*, v, v^*)$   % Eq. (3.10)
18:     $\theta \leftarrow \theta - \eta \nabla_\theta \mathcal{L}$
19:   **end for**
20: **end for**

where $L_{cls}$ is the Focal Loss for classification to handle class imbalance, $L_{reg}$ is the temporal Intersection over Union (tIoU) Loss for robust boundary regression, and $L_{center}$ is a Binary Cross-Entropy (BCE) Loss for the center-ness score. The parameters $\lambda$ and $\mu$ balance the contributions of the regression and center-ness losses. The overall training process is summarized in Algorithm 1.

## 4.5 Key Summary Generation

During inference, the trained model produces importance scores ($s_j$), segment boundary predictions ($\delta_l, \delta_r$), and centerness scores ($v_j$) for each frame $j$.

1. **Segment Proposal:** Candidate summary segments are proposed for each frame, with start and end times calculated as $t_{start} = j - \delta_t$ and $t_{end} = j + \delta_r$. A confidence score is calculated as $co_j = s_j \times v_j$.

2. **Non-Maximum Suppression (NMS):** To eliminate highly overlapping and redundant segment proposals, NMS is applied based on the confidence scores.

3. **Shot Segmentation with KTS:** The original video is segmented into coherent shots using the Kernel Temporal Segmentation (KTS) algorithm [Potapov *et al.*, 2014], an unsupervised method for detecting change points based on frame similarity.

4. **Shot Importance Scoring:** The frame-level importance scores are aggregated (e.g., by averaging) within

each KTS-defined shot to obtain a shot-level importance score $y_h$.

5. **0/1 Knapsack Selection:** The final set of shots for the summary is selected by solving a 0/1 knapsack problem. The goal is to maximize the total shot importance scores subject to a constraint on the total summary length (typically 15% of the original video's duration). The problem can be formulated as:

$$\max \sum_{h=1}^{C} u_h y_h, \quad \text{s.t.} \sum_{h=1}^{C} u_h l_h \leq 0.15 \times L_{total}$$

where $u_h \in {0,1}$ indicates if shot h is selected, $y_h$ and $l_h$ are its importance and length, $C$ is the total number of shots, and $L_{total}$ is the total video length. The selected shots are concatenated to form the final video summary.

## 5 Experiments

This section evaluates the proposed **MF2Summ** framework on benchmark video-summarisation datasets. All experiments are implemented in PYTORCH and executed on an NVIDIA RTX 4090 GPU.

### 5.1 Experimental Set-Up

**Datasets**
We evaluate our model on two widely used benchmark datasets for video summarization:

- **TVSum**: 50 videos from 10 categories collected from YouTube, each annotated by 20 users.
- **SumMe** : 25 raw user-generated videos with dense annotations.

Following standard practice, all videos are sampled at 2 fps. Audio tracks are re-sampled to 16 kHz.

**Evaluation Metrics**
Following previous work [Zhu et al., 2020; ?], we report the maximum *F-score* between machine-generated summaries and ground-truth user summaries, computed with a 15% summary length constraint.

**Implementation Details**
We initialize the visual encoder with pre-trained GoogLeNet weights [Chen et al., 2019] and the audio encoder with pre-trained SoundNet weights [Aytar et al., 2016]. The network is trained for up to 300 epochs using the ADAM optimizer ($\eta = 1e-5$, $\beta_{1,2}=0.9, 0.999$). Loss weights are fixed to $\lambda=\mu=1$. During inference, we apply the Knapsack algorithm to ensure the 15% selection budget is met. Table 1 compares MF2Summ with a broad range of single- and multi-modal state-of-the-art methods.

In SumMe, MF2Summ achieves an absolute F-score gain of **1.9 pp** over the strong visual baseline DSNet-AF and **1.5 pp** over the language-guided CLIP-It. On TVSum, MF2Summ still surpasses DSNet-AF by **0.6 pp**, while trailing CLIP-It by 0.9 pp. We attribute the slight shortfall on TVSum to the substantial audio noise that characterises many user-generated videos in this dataset, which hampers reliable acoustic feature extraction. Unlike CLIP-It—which fuses vision and text without explicit temporal alignment—MF2Summ employs a cross-modal Transformer augmented with an alignment-guided self-attention mask, enabling it to model audio–visual correlations more faithfully and exploit multi-modal cues to greater effect.

Table 1: Comparison of F-scores (%) on the **SumMe** and **TVSum** benchmarks.

| Method | SumMe | TVSum |
|---|---|---|
| vsLSTM [Zhang et al., 2016b] | 37.6 | 54.2 |
| dppLSTM [Zhang et al., 2016b] | 38.6 | 54.7 |
| SUM-GAN [Mahasseni et al., 2017] | 41.7 | 56.3 |
| DR-DSN [Zhou et al., 2018] | 42.1 | 58.1 |
| A-AVS [Ji et al., 2019] | 43.9 | 59.4 |
| M-AVS [Ji et al., 2019] | 44.4 | 61.0 |
| FCSN [Rochan et al., 2018] | 48.8 | 58.4 |
| VASNet [Fajtl et al., 2019] | 49.7 | 61.4 |
| DSNet-AB [Zhu et al., 2020] | 50.2 | 62.1 |
| DSNet-AF [Zhu et al., 2020] | 51.2 | 61.9 |
| CLIP-It [Narasimhan et al., 2021] | 51.6 | **64.2** |
| **MF2Summ (ours)** | **53.1** | 63.3 |

### 5.2 Ablation Study

To quantify the contribution of our key components, we conduct a focused ablation study. Table 2 shows that removing the audio modality or specific architectural choices significantly degrades performance, demonstrating their importance.

Table 2: Ablation results on TVSum and SumMe (*F-score*, %).

| Model Variant | TVSum | SumMe |
|---|---|---|
| Full model (MF2Summ) | **63.3** | **53.1** |
| w/o Audio (Video-Only) | 61.9 | 51.2 |
| w/o Alignment-Guidance | 61.4 | 52.2 |
| w/o $\mathcal{L}_{\text{center}}$ | 61.8 | 47.5 |

The results validate three key insights:

1. Multi-modal fusion is critical for capturing complementary semantics in audiovisual streams, with the audio modality providing a performance boost of 1.4-1.9%.

2. The alignment-guided fusion mechanism is superior to global attention, preventing performance degradation from noisy or misaligned segments.

3. The joint optimization of classification, regression, and centre-ness yields better localization than optimizing any objective in isolation. The centre-ness branch, in particular, provides a crucial confidence signal.

### 5.3 Qualitative Analysis

Figure 6 visualises generated summaries on challenging examples. MF2Summ successfully identifies key-event segments while avoiding redundant or low-impact frames, thanks to the centre-ness guidance that emphasizes middle-shot frames and the tIoU regression that refines boundary placement.
We conclude that MF2Summ sets a new state-of-the-art for audio-visual video summarisation while maintaining a conceptually clear and effective framework.

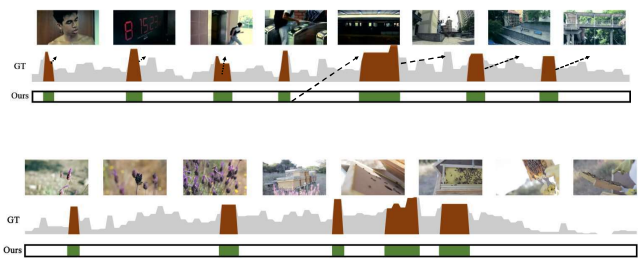

Figure 6: Qualitative comparison between ground-truth (red bar) and MF2Summ predictions (green bars).

## 6 Conclusion

This paper introduced MF2Summ, a novel video summarization model that effectively leverages multimodal content understanding by integrating visual and auditory information. To address the limitations of traditional unimodal approaches, which often fail to capture the full semantic richness of videos , our model employs a sophisticated fusion architecture. The core of MF2Summ is a dual-transformer design, featuring a cross-modal Transformer for inter-modality interaction and an alignment-guided self-attention Transformer for robust, temporally-aware fusion . This is complemented by a multi-task learning framework that predicts segment importance, boundaries, and center-ness to generate high-quality summaries .

Comprehensive experiments on the SumMe and TVSum benchmark datasets demonstrated that MF2Summ achieves highly competitive performance, outperforming strong visual-only baselines and other state-of-the-art methods. Ablation studies systematically validated the significant contributions of the auditory modality, the cross-modal interaction mechanism, and the alignment-guided fusion module to the overall performance.

Despite its promising results, MF2Summ has certain limitations. Its performance can be sensitive to the quality of the audio input, as some videos with noisy audio tracks may hinder feature extraction. Future research could explore several exciting directions. First, incorporating additional modalities, such as textual information from automatic speech recognition (ASR) or optical character recognition (OCR), could provide a more holistic understanding of video content . Second, investigating end-to-end training of the feature extractors along with the summarization network might allow for more task-specific feature learning. Finally, developing techniques to improve the model's robustness to noisy audio would enhance its applicability to challenging real-world videos. In summary, MF2Summ demonstrates that a well-designed audio-visual fusion strategy can significantly advance the state of the art in video summarization.

## References

[Aytar et al., 2016] Y. Aytar, C. Vondrick, and A. Torralba. Soundnet: Learning sound representations from unlabeled video. In *Advances in neural information processing systems*, volume 29, 2016.

[Baltrušaitis et al., 2018] T. Baltrušaitis, C. Ahuja, and L. P. Morency. Multimodal machine learning: A survey and taxonomy. *IEEE transactions on pattern analysis and machine intelligence*, 41(2):423–443, 2018.

[Chen et al., 2019] Y. Chen, L. Tao, X. Wang, and et al. Weakly supervised video summarization by hierarchical reinforcement learning. In *Proceedings of the ACM Multimedia Asia*, pages 1–6, 2019.

[Fajtl et al., 2019] J. Fajtl, H. S. Sokeh, V. Argyriou, and et al. Summarizing videos with attention. In *Computer Vision–ACCV 2018 Workshops*, pages 39–54, 2019.

[Jangra et al., 2023] A. Jangra, S. Mukherjee, A. Jatowt, and et al. A survey on multi-modal summarization. *ACM Computing Surveys*, 55(13s):1–36, 2023.

[Ji et al., 2019] Z. Ji, K. Xiong, Y. Pang, and et al. Video summarization with attention-based encoder–decoder networks. *IEEE Transactions on Circuits and Systems for Video Technology*, 30(6):1709–1717, 2019.

[Mahasseni et al., 2017] B. Mahasseni, M. Lam, and S. Todorovic. Unsupervised video summarization with adversarial lstm networks. In *Proceedings of the IEEE conference on Computer Vision and Pattern Recognition*, pages 202–211, 2017.

[Narasimhan et al., 2021] M. Narasimhan, A. Rohrbach, and T. Darrell. Clip-it! language-guided video summarization. In *Advances in Neural Information Processing Systems*, volume 34, pages 13988–14000, 2021.

[Panda et al., 2017] R. Panda, A. Das, Z. Wu, and et al. Weakly supervised summarization of web videos. In *Proceedings of the IEEE international conference on computer vision*, pages 3657–3666, 2017.

[Potapov et al., 2014] D. Potapov, M. Douze, Z. Harchaoui, and et al. Category-specific video summarization. In *Computer Vision–ECCV 2014: 13th European Conference, Zurich, Switzerland, September 6-12, 2014, Proceedings, Part VI 13*, pages 540–555. Springer International Publishing, 2014.

[Psallidas et al., 2021] T. Psallidas, P. Koromilas, T. Giannakopoulos, and et al. Multimodal summarization of user-generated videos. *Applied Sciences*, 11(11):5260, 2021.

[Rochan et al., 2018] M. Rochan, L. Ye, and Y. Wang. Video summarization using fully convolutional sequence networks. In *Proceedings of the European conference on computer vision (ECCV)*, pages 347–363, 2018.

[Szegedy et al., 2015] C. Szegedy, W. Liu, Y. Jia, and et al. Going deeper with convolutions. In *Proceedings of the IEEE conference on computer vision and pattern recognition*, pages 1–9, 2015.

[Tsai *et al.*, 2019] Y. H. H. Tsai, S. Bai, P. P. Liang, and et al. Multimodal transformer for unaligned multimodal language sequences. In *Proceedings of the conference. Association for computational linguistics. Meeting*, page 6558, 2019.

[Vaswani *et al.*, 2017] Ashish Vaswani, Noam Shazeer, Niki Parmar, Jakob Uszkoreit, Llion Jones, Aidan N. Gomez, Łukasz Kaiser, and Illia Polosukhin. Attention is all you need. In *Advances in Neural Information Processing Systems*, volume 30, pages 5998–6008, 2017.

[Zhang *et al.*, 2016a] K. Zhang, W. L. Chao, F. Sha, and et al. Video summarization with long short-term memory. In *Computer Vision–ECCV 2016: 14th European Conference, Amsterdam, The Netherlands, October 11–14, 2016, Proceedings, Part VII 14*, pages 766–782. Springer International Publishing, 2016.

[Zhang *et al.*, 2016b] K. Zhang, W. L. Chao, F. Sha, and et al. Video summarization with long short-term memory. In *Computer Vision–ECCV 2016: 14th European Conference, Amsterdam, The Netherlands, October 11–14, 2016, Proceedings, Part VII 14*, pages 766–782. Springer International Publishing, 2016.

[Zhou *et al.*, 2018] K. Zhou, Y. Qiao, and T. Xiang. Deep reinforcement learning for unsupervised video summarization with diversity-representativeness reward. In *Proceedings of the AAAI conference on artificial intelligence*, volume 32, 2018.

[Zhu *et al.*, 2020] W. Zhu, J. Lu, J. Li, and et al. DSNet: A flexible detect-to-summarize network for video summarization. *IEEE Transactions on Image Processing*, 30:948–962, 2020.